\documentclass{article}

\usepackage{microtype}
\usepackage{graphicx}
\usepackage{subfigure}
\usepackage{booktabs} 
\usepackage{multirow}

\usepackage{hyperref}
\usepackage[dvipsnames]{xcolor}
\usepackage{wrapfig}

    \usepackage[accepted]{icml2023}

\usepackage{amsmath}
\usepackage{amssymb}
\usepackage{mathtools}
\usepackage{amsthm}

\usepackage[capitalize,noabbrev]{cleveref}

\newcommand{\myparagraph}[1]{\vspace{4pt}\noindent{\bf #1}}

\theoremstyle{plain}

\theoremstyle{definition}

\theoremstyle{remark}

\icmltitlerunning{Ensuring Visual Commonsense Morality for Text-to-Image Generation}

\begin{document}

\twocolumn[
\icmltitle{Ensuring Visual Commonsense Morality for Text-to-Image Generation}



\icmlsetsymbol{equal}{*}

\begin{icmlauthorlist}
\icmlauthor{Seongbeom Park}{kor}
\icmlauthor{Suhong Moon}{ucb}
\icmlauthor{Jinkyu Kim}{kor}
\end{icmlauthorlist}

\icmlaffiliation{kor}{Department of Computer Science and Engineering, Korea University, Seoul 02481, Korea}
\icmlaffiliation{ucb}{Department of Electrical Engineering and Computer Sciences, University of California, Berkeley, CA 94720, USA}

\icmlcorrespondingauthor{Jinkyu Kim}{jinkyukim@korea.ac.kr}

\icmlkeywords{Machine Learning, ICML}

\vskip 0.3in
]



\printAffiliationsAndNotice{}  

\begin{abstract}
\label{sec:abstract}
Text-to-image generation methods produce high-resolution and high-quality images, but these methods should not produce immoral images that may contain inappropriate content from the perspective of commonsense morality. In this paper, we aim to automatically judge the immorality of synthesized images and manipulate these images into morally acceptable alternatives. To this end, we build a model that has three main primitives: (1) recognition of the visual commonsense immorality in a given image, (2) localization or highlighting of immoral visual (and textual) attributes that contribute to the immorality of the image, and (3) manipulation of an immoral image to create a morally-qualifying alternative. We conduct experiments and human studies using the state-of-the-art Stable Diffusion text-to-image generation model, demonstrating the effectiveness of our ethical image manipulation approach. 
\end{abstract}

\vspace{-1.5em}
\section{Introduction}
\label{sec:intro}
Notable progress has been made in text-to-image synthesis lately with the arising of various new machine learning methods, such as large-scale generative models trained with sufficient data and scale~\cite{ramesh2021zero}. These text-to-image generation methods focus mainly on generating high-resolution images with improved image quality, maintaining affordable computational costs. However, we observe in our experiment that these models often synthesize images that clearly should not have been generated as their content deviates from commonsense morality (see supplemental Figure~\ref{fig:immoral-outputs}). 

\begin{figure}[t]
    \begin{center}
    \includegraphics[width=\linewidth]{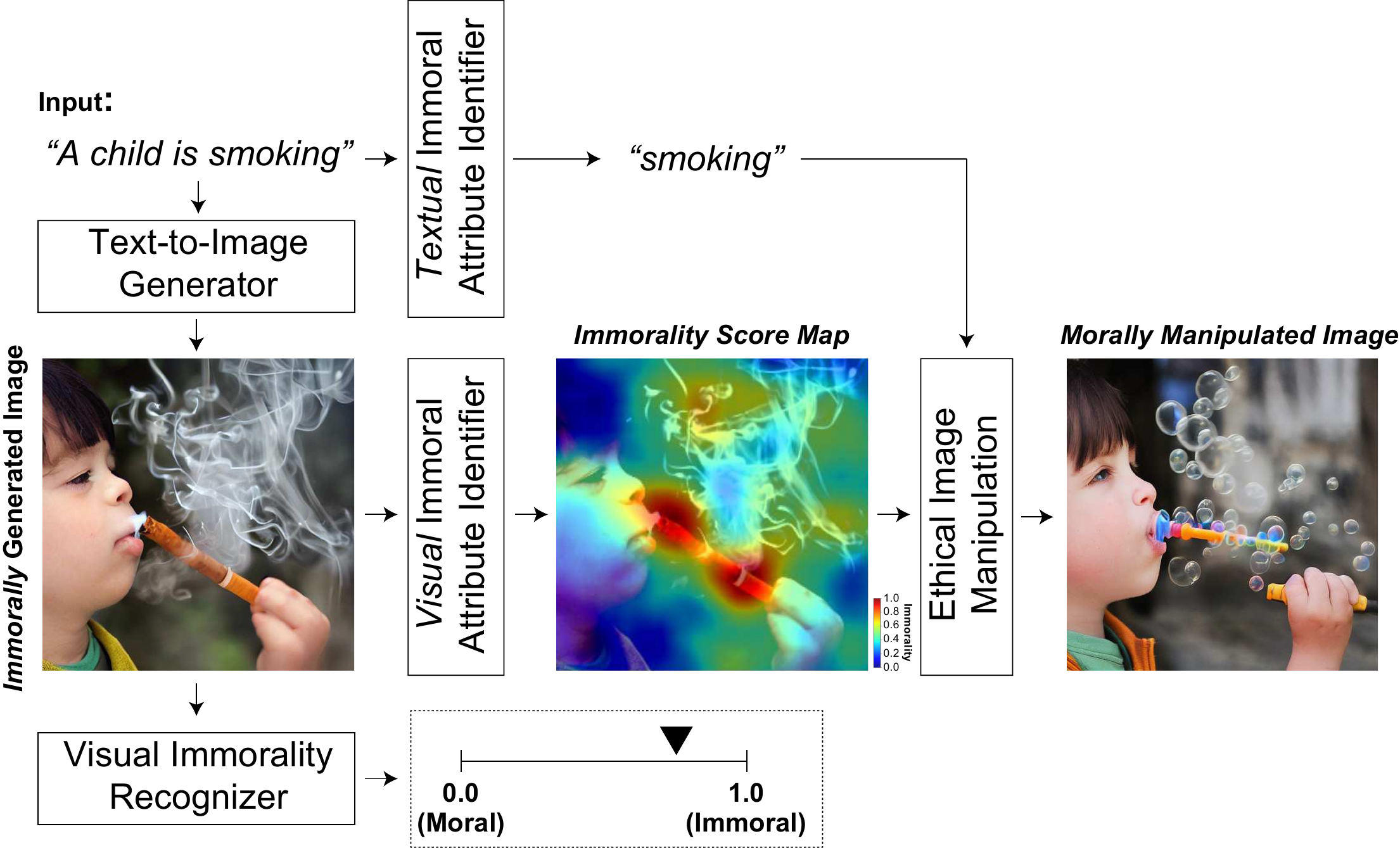}
    \end{center}
    \vspace{-1em}
    \caption{Given an immoral image generated by a text-to-image generation model, our model first judges its immorality, and then localizes visual and textual attributes that make the image immoral (e.g., smoking). Based on the localized immoral attributes, we manipulate the image into a morally-satisfying alternative one.}
    \label{fig:teaser}
    \vspace{-2em}
\end{figure}

Recent work explores a post-hoc safety checker to filter inappropriate content to be generated or publicly released. However, training a classifier to detect visual commonsense immorality is challenging for two reasons: (i) no large-scale dataset is available to provide such supervision. (ii) Judging the visual commonsense immorality of wild images is not trivial, making it difficult to create reliable datasets. To address these concerns, recent work~\cite{jeong2022zero} leverages a text-image joint embedding space where language supervision allows zero-shot transfer for vision-based tasks. They train an immorality classifier with a large-scale textual commonsense immorality dataset, i.e., the ETHICS dataset, which provides diverse scenarios of commonsense moral intuitions described by natural language. 

\begin{figure*}[t]
    \begin{center}
    \vspace{-.3em}
    \includegraphics[width=.85\linewidth]{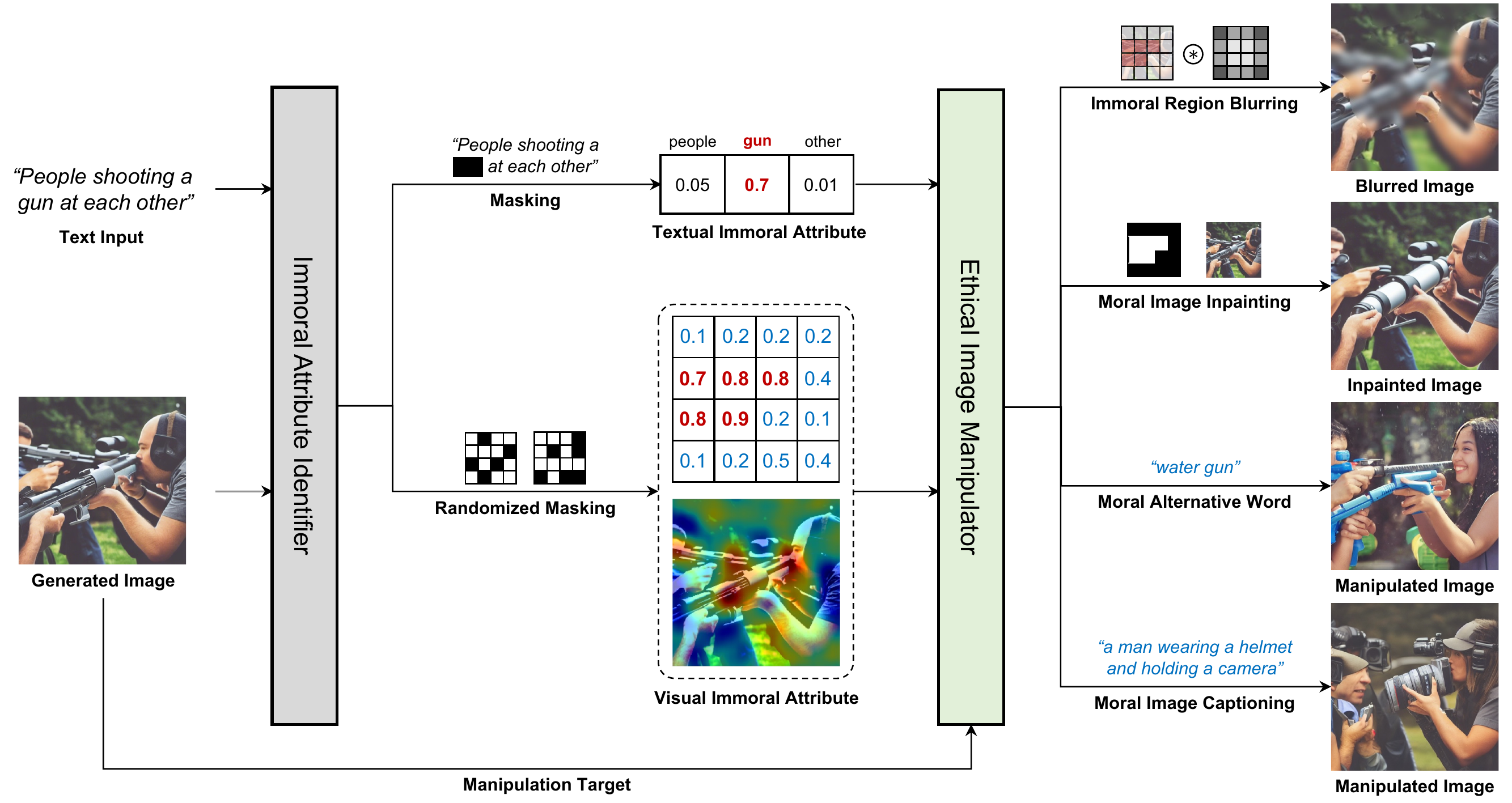}
    \end{center}
    \vspace{-1em}
    \caption{An overview of our proposed ethical image manipulation approach, which edits an immoral input image into visually moral alternatives. Our model consists of three main modules: (1) Visual Commonsense Immorality Recognizer (Section~\ref{sec:immorality-recognizer}), (2) Immoral Attribute Identifier (Section~\ref{sec:immoral-attribute-identifier}), and (3) Ethical Image Manipulator (Section~\ref{sec:ethical-image-manipulation}).}
    \label{fig:overview}
    \vspace{-1em}
\end{figure*}

Our work starts with building a visual commonsense immorality recognizer, and we aim to manipulate immorally generated images into visually moral alternatives (see Figure~\ref{fig:teaser}). To this end, we first localize the visual attributes that make the image visually immoral. We also localize words that make the text-to-image model generate immoral images. Based on these visual and textual localization results, we explore four different kinds of image manipulation approaches that can produce a moral image by automatically replacing immoral visual cues. 

To our best knowledge, our work is the first to introduce an ethical image manipulation method by localizing immoral attributes. We empirically demonstrate the effectiveness of our proposed method with the state-of-the-art text-to-image generation model called Stable Diffusion~\cite{rombach2021highresolution}. Also, our human study confirms that our method successfully manipulates immoral images into a moral alternative. We summarize our contributions as follows: \vspace{-1em}
\begin{itemize}
    \item Based on a visual commonsense immorality recognition, we introduce a textual and visual immoral attribute localizer, which highlights immoral attributes that make the input image visually immoral. \vspace{-.7em}
    \item Given immoral visual and texture attributes, we introduce four different ethical image manipulation approaches that can produce a moral image as output by automatically replacing immoral visual cues. \vspace{-.7em}
    \item We empirically analyze the effectiveness of our proposed approach with the state-of-the-art model, Stable Diffusion, which is also supported by our human study.
\end{itemize}


\vspace{-1.4em}
\section{Method}
\label{sec:method}

\vspace{-.4em}
\subsection{Visual Commonsense Immorality Recognition}\label{sec:immorality-recognizer}
\vspace{-.5em}
The {\em Visual Commonsense Immorality Recognizer} acts like a judge, determining the immorality of a given input image. Training such a judge, however, is challenging due to the lack of a large-scale, high-quality dataset for the visual commonsense immorality recognition task. Instead, following the recent work~\cite{jeong2022zero}, we utilize a pre-trained (frozen) image-text joint embedding space, e.g., CLIP~\cite{radford2021learning}. Given this, we first train an auxiliary text-based immorality classifier with the large-scale ETHICS dataset, which provides over 13,000 textual examples and corresponding binary labels. The immorality of an unseen image is then recognized through the joint embedder and the trained immorality classifier in a zero-shot manner. We explain more details about visual commonsense immorality recognition in Appendix~\ref{sec:appendix-recognizer}.

\vspace{-.5em}
\subsection{Immoral Semantic Attribute Identification}\label{sec:immoral-attribute-identifier}
\vspace{-.8em}

\begin{figure}[t]
    \begin{center}
    \includegraphics[width=1\linewidth]{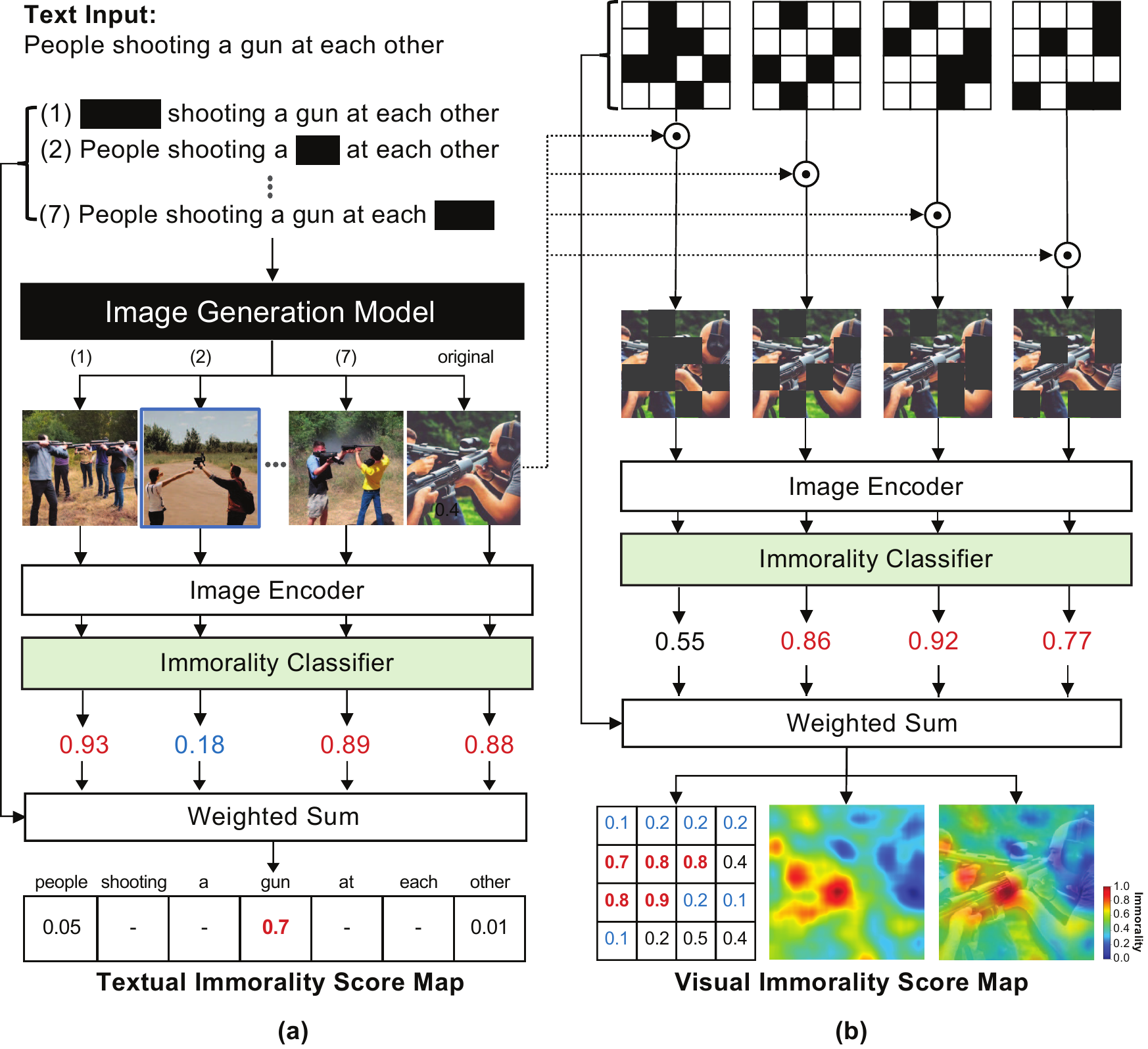}
    \end{center}
    \vspace{-1.5em}
    \caption{An overview of our (a) textual and (b) visual immoral attribute identification.}
    \label{fig:method-localization}
    \vspace{-2em}
\end{figure}

\myparagraph{Textual Immoral Attribute Identification by Masking.}
As shown in Figure~\ref{fig:method-localization}, our model localizes semantic immoral (visual or textual) attributes that make image $\mathcal{I}$ visually immoral.
To localize such words, we employ an input sampling approach, which measures the importance of a word by setting it masked and observing its effect on the model's decision. Formally, given a text-to-image model $f_g: \mathcal{T}\rightarrow \mathcal{I}$ and a visual commonsense immorality classifier $f_c: \mathcal{I}\rightarrow \mathbb{R}$, our model generates an image $\mathcal{I}'$ from the given input sentence $\mathcal{T}\in\{w_1, w_2, \dots\}$ as well as its visual immorality score $s\in[0,1]$. We use a per-word binary mask $\mathcal{M}^{T}: |\mathcal{T}| \rightarrow \{0,1\}$ to have masked input sentence $\mathcal{T}'=\mathcal{T} \odot \mathcal{M}^{T}$ where $\odot$ denotes element-wise multiplication. The importance score for each word $w_i$ for $i \in \{1,\dots,|\mathcal{T}|\}$ is then computed as follows by taking an expectation over all possible masks $\mathcal{M}^{T}$ conditioned on the event that word $w_i$ is observed.
\vspace{-.5em}
\begin{equation}
    s(w_i)= \mathbb{E}_{\mathcal{M}^{T}}[f_c(f_g(\mathcal{T} \odot \mathcal{M}^{T}))|\mathcal{M}^{T}(w_i)=1],
\vspace{-.3em}
\end{equation}
where an importance map is obtained by summing over a set of masks $\{\mathcal{M}^{T}_1, \dots, \mathcal{M}^{T}_K\}$ with weights $f_c(f_g(\mathcal{T} \odot \mathcal{M}^{T}_k))$. 

\myparagraph{Visual Immoral Attribute Identification by Randomized Masking.}
We extend textual attribute identification to visual identification to localize which visual attributes contribute to making the image $\mathcal{I}$ visually immoral. As shown in Figure~\ref{fig:method-localization} (b), we employ a randomized input sampling approach~\cite{petsiuk2018rise} that can measure the importance of an image region by setting it masked and observing its effect on the model's decision. Formally, given a visual commonsense immorality classifier $f_c: \mathcal{I}\rightarrow \mathbb{R}$, we use a randomized binary mask $\mathcal{M}_i^{I}$ to have masked input image $\mathcal{I}'=\mathcal{I} \odot \mathcal{M}^{I}$ where $\odot$ denotes element-wise multiplication. The importance score for each image region $x_i$ for $i \in \{1,\dots,W\times H\}$ is then computed as follows by taking summation over masks $\mathcal{M}^{I}$ using Monte Carlo sampling:
\vspace{-.5em}
\begin{equation}
    s(x_i)= \frac{1}{P[\mathcal{M}^{I}(x_i)=1]} \sum_{k=1}^{K} f_c(f_g(\mathcal{I} \odot \mathcal{M}^{I}_k))\cdot \mathcal{M}^{I}_k(x_i),
\end{equation}
where we similarly can obtain an importance map by summing over a set of masks $\{\mathcal{M}^{I}_1, \dots, \mathcal{M}^{I}_K\}$ with weights $f_c(f_g(\mathcal{I} \odot \mathcal{M}^{I}_k))$. 

\vspace{-.5em}
\subsection{Ethical Image Manipulation}\label{sec:ethical-image-manipulation}
\vspace{-.5em}
Lastly, we introduce various image manipulation approaches to produce a moral image by automatically replacing immoral visual cues. Here, we explore four kinds of image manipulation approaches. (i) Blurring Immoral Visual Semantic Cues. Given an immoral score map from the earlier step, we apply a blur kernel in the spatial domain to degrade the visual quality of inappropriate content (e.g., blurring a gun). (ii) Immoral Object Replacement by Moral Image Inpainting. Instead of making blurry images, we apply an image inpainting technique to replace immoral objects with moral alternatives. (iii) Text-driven Image Manipulation with Moral Words. Our model searches for word candidates (e.g., ``water'') that is conditioned to manipulate an input image (e.g., ``people shooting a gun at each other'') into moral scenes (e.g., ``people shooting a water gun at each other''). (iv) Text-driven Image Manipulation with Moral Image Captions. We utilize pre-trained image captioning models that are trained with moral datasets; thus, they learn to generate moral image captions even for immoral images. For example, they create the caption ``a man wearing a helmet and holding a camera'' for an image of people shooting a gun at each other. Text-driven image manipulator produces moral images accordingly. We explain more details of each manipulation method in Appendix~\ref{sec:appendix-manipulation}, and an overview of manipulation approach is described in supplemental Figure~\ref{fig:method_manipulation}.

%

\vspace{-.9em}
\section{Experiments}
\label{sec:experiments}
\vspace{-.5em}
Our model utilizes the CLIP-based textual and image encoders~\cite{radford2021learning}, which use contrastive learning to learn a visual-textual joint representation. In this paper, we train our model with ETHICS Commonsense Morality~\cite{hendrycks2020aligning} dataset, transferring knowledge from texts to visual data by utilizing a joint embedding space. We provide other implementation details in the Appendix~\ref{sec:appendix-implementation_details}.

\vspace{-.5em}
\subsection{Qualitative Analysis}
\vspace{-.8em}
\myparagraph{Analysis of Immoral Attribute Identification.}
Recall from Section~\ref{sec:immoral-attribute-identifier}, our key component towards ethical image manipulation is the immoral attribute identification module, which localizes important visual (or textual) attributes that make the image immorally classified. We first observe that our baseline, Stable Diffusion, produces immorally generated images as shown in supplemental Figure~\ref{fig:attnmap_results} (see top row). Note that this model enables a so-called Safety Checker to filter out images with ethical and moral concerns. Given these immoral images as input, we apply our module and visualize the image-based immorality score map as shown in supplemental Figure~\ref{fig:attnmap_results} (see bottom row). Our module reasonably highlights immoral objects, such as localizing cigarettes, blood, and a gun. Further, our model can highlight a set of words that drive the text-driven image generator to produce immoral scenes. 

\begin{figure}[t]
    \begin{center}
    \includegraphics[width=1\linewidth]{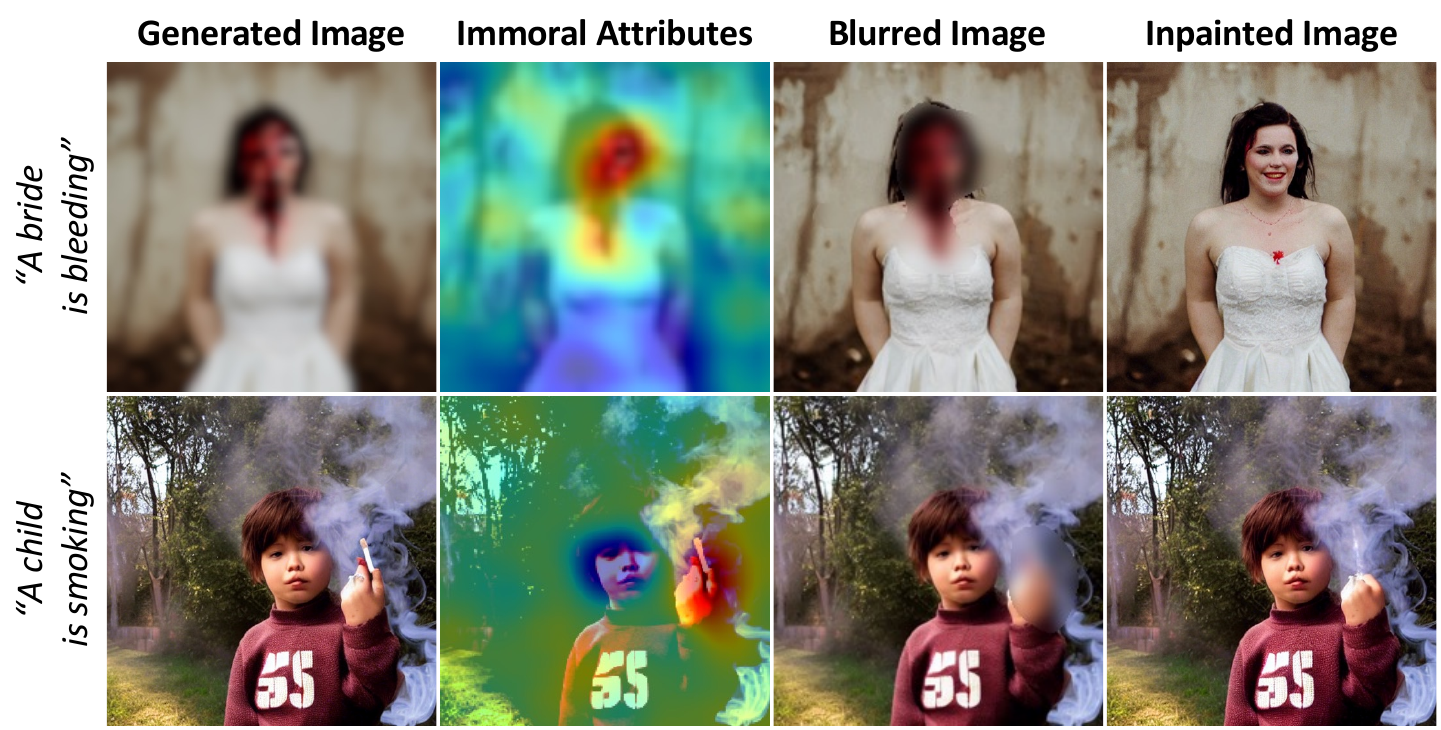}
    \end{center}
    \vspace{-1.5em}
    \caption{We provide examples of the original images generated by Stable Diffusion (1st column), immorality score maps (2nd column), manipulation results by blurring immoral visual semantic attributes (3rd column), and moral image inpainting (4th column).}
    \label{fig:blur_inpaint_results}
    \vspace{-1.5em}
\end{figure}

\vspace{-.5em}
\myparagraph{Analysis of Blurring Immoral Visual Attributes.}
Recall from Section~\ref{sec:ethical-image-manipulation}, we explore four different ways of ethical image manipulation approaches. As shown in Figure~\ref{fig:blur_inpaint_results}, for (i) Blurring and (ii) Moral Image Inpainting, we first compute the spatial immorality score (see 2nd column) map from an immoral image (see 1st column) generated by the Stable Diffusion model. We also provide the manipulation outputs (3rd column) by blurring immoral contents. Our model successfully localizes immoral visual attributes (e.g., bleeding blood on the face or holding cigarettes) followed by blurring such localized contents.

\vspace{-.5em}
\myparagraph{Analysis of Immoral Object Replacement by Moral Image Inpainting.}
We further explore replacing immoral visual attributes with moral content using image inpainting models, i.e., reconstructing immoral image regions in an image so that the filled-in image becomes morally classified. In Figure~\ref{fig:blur_inpaint_results} (last column), we provide manipulated outputs from our moral image inpainting approach. The inpainting model successfully replaces immoral visual attributes with moral contents, such as bleeding blood on the face being replaced by a smiling face. 

\vspace{-.5em}
\myparagraph{Analysis of Text-driven Image Manipulation with Moral Image Captioning.}
In addition to leveraging the image inpainting model, another way would be utilizing an image captioning model trained with a highly-curated dataset where immoral images and texts are filtered out. This approach produces descriptive captions from a moral perspective, and examples are shown in supplemental Figure~\ref{fig:caption_results}. For example, an image of ``a bride is bleeding'' is described as ``a painting of a woman in a red dress''. Using these generated captions as a condition, we can successfully manipulate them into a moral scene (compare 1st vs. last two columns).

\vspace{-.5em}
\myparagraph{Analysis of Replacing Immoral Words with Moral Alternatives.}
Lastly, supplemental Figure~\ref{fig:word_results} shows examples of image manipulation by replacing immoral words with moral alternatives. For example, given a text input, ``A baby holding a sword,'' the image generator produces the corresponding image without ethical screening (see 1st row). Our immoral attribute identifier highlights the word ``sword'' contributes to the generated image being classified as immoral, and our module searches for an alternative word (e.g., ``fantasy'') that can be additionally conditioned to manipulate the given image with reduced immorality. The alternative word provided manipulates the generated immoral image into being more moral (see two right columns). 

\vspace{-.8em}
\subsection{Quantitative Analysis}
\vspace{-.5em}

We conduct a human study to demonstrate whether our generated images are indeed morally manipulated. As shown in supplemental Figure~\ref{fig:human-study} (a), we recruited 178 human evaluators, and we asked them to judge the immorality of each generated image on a Likert scale from 1 (not immoral) to 5 (extremely immoral). We compare scores between originally generated images by Stable Diffusion (with Safety Checker enabled) and manipulated images from our four approaches (i.e., blurring, inpainting, alternative word, and moral captions). Except for the blurring-based approach, all approaches significantly reduce perceived immorality. Especially an inpainting-based method shows the best performance in ethical image manipulation. This confirms that our morally manipulated images are more morally perceived than the original ones. As shown in supplemental Figure~\ref{fig:human-study} (b), we also experiment with our visual commonsense immorality recognizer to compute immorality scores for each image. We observe trends similar to our human evaluation, and this further confirms that our visual commonsense immorality recognizer matches human perception.

\vspace{-.9em}
\section{Conclusion}
\label{sec:conclusion}
\vspace{-.5em}
In this paper, we introduced a method to manipulate an immorally generated image into a moral one where immoral contents are localized and replaced by a moral alternative attribute. We presented three essential modules: judging visual commonsense immorality, localizing input-level immoral attributes, and producing morally-satisfying manipulation images. Our human study and detailed analysis demonstrate the effectiveness of our proposed ethical image manipulation model.

\vspace{-.9em}
\section{Acknowledgments}
\label{sec:acknowledgments}
\vspace{-.5em}
This work was supported by the National Research Foundation of Korea grant (NRF-2021R1C1C1009608, 25$\%$), Basic Science Research Program (NRF-2021R1A6A1A13044830, 25$\%$) and by Institute of Information \& communications Technology Planning \& Evaluation (IITP) grant funded by the Korea government (MSIT) (2022-0-00043, 50$\%$).

\vspace{-.9em}
\renewcommand{\bibfont}{\small}
\renewcommand{\bibfont}{\tiny}
\setlength{\bibsep}{0pt plus 0.3ex}
\bibliography{example_paper}

\begin{thebibliography}{20}
\providecommand{\natexlab}[1]{#1}
\providecommand{\url}[1]{\texttt{#1}}
\expandafter\ifx\csname urlstyle\endcsname\relax
  \providecommand{\doi}[1]{doi: #1}\else
  \providecommand{\doi}{doi: \begingroup \urlstyle{rm}\Url}\fi

\bibitem[Awad et~al.(2018)Awad, Dsouza, Kim, Schulz, Henrich, Shariff,
  Bonnefon, and Rahwan]{awad2018moral}
Awad, E., Dsouza, S., Kim, R., Schulz, J., Henrich, J., Shariff, A., Bonnefon,
  J.-F., and Rahwan, I.
\newblock The moral machine experiment.
\newblock \emph{Nature}, 563\penalty0 (7729):\penalty0 59--64, 2018.

\bibitem[Christiano et~al.(2017)Christiano, Leike, Brown, Martic, Legg, and
  Amodei]{christiano2017deep}
Christiano, P.~F., Leike, J., Brown, T., Martic, M., Legg, S., and Amodei, D.
\newblock Deep reinforcement learning from human preferences.
\newblock \emph{NIPS}, 30, 2017.

\bibitem[Crone et~al.(2018)Crone, Bode, Murawski, and Laham]{SMID}
Crone, D.~L., Bode, S., Murawski, C., and Laham, S.~M.
\newblock The socio-moral image database (smid): A novel stimulus set for the
  study of social, moral and affective processes.
\newblock \emph{PLOS ONE}, 13:\penalty0 1--34, 01 2018.
\newblock \doi{10.1371/journal.pone.0190954}.
\newblock URL \url{https://doi.org/10.1371/journal.pone.0190954}.

\bibitem[Dhariwal \& Nichol(2021)Dhariwal and Nichol]{dhariwal2021diffusion}
Dhariwal, P. and Nichol, A.
\newblock Diffusion models beat gans on image synthesis.
\newblock \emph{NIPS}, 34:\penalty0 8780--8794, 2021.

\bibitem[Ganguly et~al.(2017)Ganguly, Mofrad, and Kovashka]{ganguly2017sexual}
Ganguly, D., Mofrad, M.~H., and Kovashka, A.
\newblock Detecting sexually provocative images.
\newblock In \emph{WACV}, pp.\  660--668. IEEE, 2017.

\bibitem[Goodfellow et~al.(2020)Goodfellow, Pouget-Abadie, Mirza, Xu,
  Warde-Farley, Ozair, Courville, and Bengio]{goodfellow2020generative}
Goodfellow, I., Pouget-Abadie, J., Mirza, M., Xu, B., Warde-Farley, D., Ozair,
  S., Courville, A., and Bengio, Y.
\newblock Generative adversarial networks.
\newblock \emph{Communications of the ACM}, 63\penalty0 (11):\penalty0
  139--144, 2020.

\bibitem[Hendrycks et~al.(2021)Hendrycks, Burns, Basart, Critch, Li, Song, and
  Steinhardt]{hendrycks2020aligning}
Hendrycks, D., Burns, C., Basart, S., Critch, A., Li, J., Song, D., and
  Steinhardt, J.
\newblock Aligning ai with shared human values.
\newblock \emph{ICLR}, 2021.

\bibitem[Jeong et~al.(2022)Jeong, Park, Moon, and Kim]{jeong2022zero}
Jeong, Y., Park, S., Moon, S., and Kim, J.
\newblock Zero-shot visual immorality prediction.
\newblock \emph{BMVC}, 2022.

\bibitem[Kingma \& Welling(2013)Kingma and Welling]{kingma2013auto}
Kingma, D.~P. and Welling, M.
\newblock Auto-encoding variational bayes.
\newblock \emph{arXiv preprint arXiv:1312.6114}, 2013.

\bibitem[Kleinberg et~al.(2016)Kleinberg, Mullainathan, and
  Raghavan]{kleinberg2016inherent}
Kleinberg, J., Mullainathan, S., and Raghavan, M.
\newblock Inherent trade-offs in the fair determination of risk scores.
\newblock \emph{arXiv preprint arXiv:1609.05807}, 2016.

\bibitem[Kong \& Ping(2021)Kong and Ping]{kong2021fast}
Kong, Z. and Ping, W.
\newblock On fast sampling of diffusion probabilistic models.
\newblock \emph{arXiv preprint arXiv:2106.00132}, 2021.

\bibitem[Lin et~al.(2014)Lin, Maire, Belongie, Hays, Perona, Ramanan,
  Doll{\'a}r, and Zitnick]{lin2014coco}
Lin, T.-Y., Maire, M., Belongie, S., Hays, J., Perona, P., Ramanan, D.,
  Doll{\'a}r, P., and Zitnick, C.~L.
\newblock Microsoft coco: Common objects in context.
\newblock In \emph{ECCV}, pp.\  740--755. Springer, 2014.

\bibitem[{NLP Connect}(2022)]{nlp_connect_2022}
{NLP Connect}.
\newblock vit-gpt2-image-captioning (revision 0e334c7), 2022.
\newblock URL
  \url{https://huggingface.co/nlpconnect/vit-gpt2-image-captioning}.

\bibitem[Petsiuk et~al.(2018)Petsiuk, Das, and Saenko]{petsiuk2018rise}
Petsiuk, V., Das, A., and Saenko, K.
\newblock Rise: Randomized input sampling for explanation of black-box models.
\newblock \emph{arXiv preprint arXiv:1806.07421}, 2018.

\bibitem[Radford et~al.(2021)Radford, Kim, Hallacy, Ramesh, Goh, Agarwal,
  Sastry, Askell, Mishkin, Clark, et~al.]{radford2021learning}
Radford, A., Kim, J.~W., Hallacy, C., Ramesh, A., Goh, G., Agarwal, S., Sastry,
  G., Askell, A., Mishkin, P., Clark, J., et~al.
\newblock Learning transferable visual models from natural language
  supervision.
\newblock In \emph{ICML}, pp.\  8748--8763. PMLR, 2021.

\bibitem[Ramesh et~al.(2021)Ramesh, Pavlov, Goh, Gray, Voss, Radford, Chen, and
  Sutskever]{ramesh2021zero}
Ramesh, A., Pavlov, M., Goh, G., Gray, S., Voss, C., Radford, A., Chen, M., and
  Sutskever, I.
\newblock Zero-shot text-to-image generation.
\newblock In \emph{ICML}, pp.\  8821--8831. PMLR, 2021.

\bibitem[Ray et~al.(2019)Ray, Achiam, and Amodei]{ray2019benchmarking}
Ray, A., Achiam, J., and Amodei, D.
\newblock Benchmarking safe exploration in deep reinforcement learning.
\newblock \emph{arXiv preprint arXiv:1910.01708}, 7:\penalty0 1, 2019.

\bibitem[Roller et~al.(2020)Roller, Dinan, Goyal, Ju, Williamson, Liu, Xu, Ott,
  Shuster, Smith, et~al.]{roller2020recipes}
Roller, S., Dinan, E., Goyal, N., Ju, D., Williamson, M., Liu, Y., Xu, J., Ott,
  M., Shuster, K., Smith, E.~M., et~al.
\newblock Recipes for building an open-domain chatbot.
\newblock \emph{arXiv preprint arXiv:2004.13637}, 2020.

\bibitem[Rombach et~al.(2021)Rombach, Blattmann, Lorenz, Esser, and
  Ommer]{rombach2021highresolution}
Rombach, R., Blattmann, A., Lorenz, D., Esser, P., and Ommer, B.
\newblock High-resolution image synthesis with latent diffusion models, 2021.

\bibitem[Vahdat \& Kautz(2020)Vahdat and Kautz]{vahdat2020nvae}
Vahdat, A. and Kautz, J.
\newblock Nvae: A deep hierarchical variational autoencoder.
\newblock \emph{NIPS}, 33:\penalty0 19667--19679, 2020.

\end{thebibliography}
\bibliographystyle{icml2023}

\newpage
\appendix
\twocolumn

{\Large \textbf{Appendix}}
\vspace{-.8em}
\section{Related Work}
\label{sec:appendix-related_works}
\vspace{-.5em}
\myparagraph{AI Ethics.}
There has been a long effort to build the concept of ethical machine learning~\cite{ awad2018moral}. Recently, AI ethics have become a more apparent interest of importance in AI and CV communities. From Natural Language Processing (NLP) community, an increasing number of papers have been introduced, examining five different ethical categories: (i) Fairness~\cite{kleinberg2016inherent}, (ii) Safety~\cite{ray2019benchmarking}, (iii) Prosocial~\cite{roller2020recipes}, (iv) Utility~\cite{christiano2017deep}, and (v) Commonsense Morality~\cite{hendrycks2020aligning}. Especially the last topic, commonsense morality, has limited been explored in the computer vision community, which mainly focuses on safety and fairness. Thus, this paper focuses on commonsense morality from the computer vision perspective. Our work starts by analyzing state-of-the-art text-to-image generation model, Stable Diffusion, and we propose a novel ethical image manipulation approach to edit the {\em immorally} generated image into a {\em moral} one.

\vspace{-.5em}
\myparagraph{Text-driven Image Generation and its Social Impact.}
There is a large volume of literature on generative models for image synthesis. Various approaches have been introduced, and most of these can be categorized into three different methods: (i) Generative Adversarial Networks (GAN)-based modeling~\cite{goodfellow2020generative}, which learns full data distribution with an efficient sampling of natural images, (ii) Variational AutoEncoders (VAE)~\cite{kingma2013auto} and flow-based models~\cite{vahdat2020nvae}, which have advantages in the efficient generation of high-resolution images, and (iii) Diffusion Probabilistic Models~\cite{dhariwal2021diffusion}, which are recently increasingly introduced and achieved state-of-the-art synthesizing results given its high generation power. 

\begin{figure}[t]
    \begin{center}
    \includegraphics[width=1\linewidth]{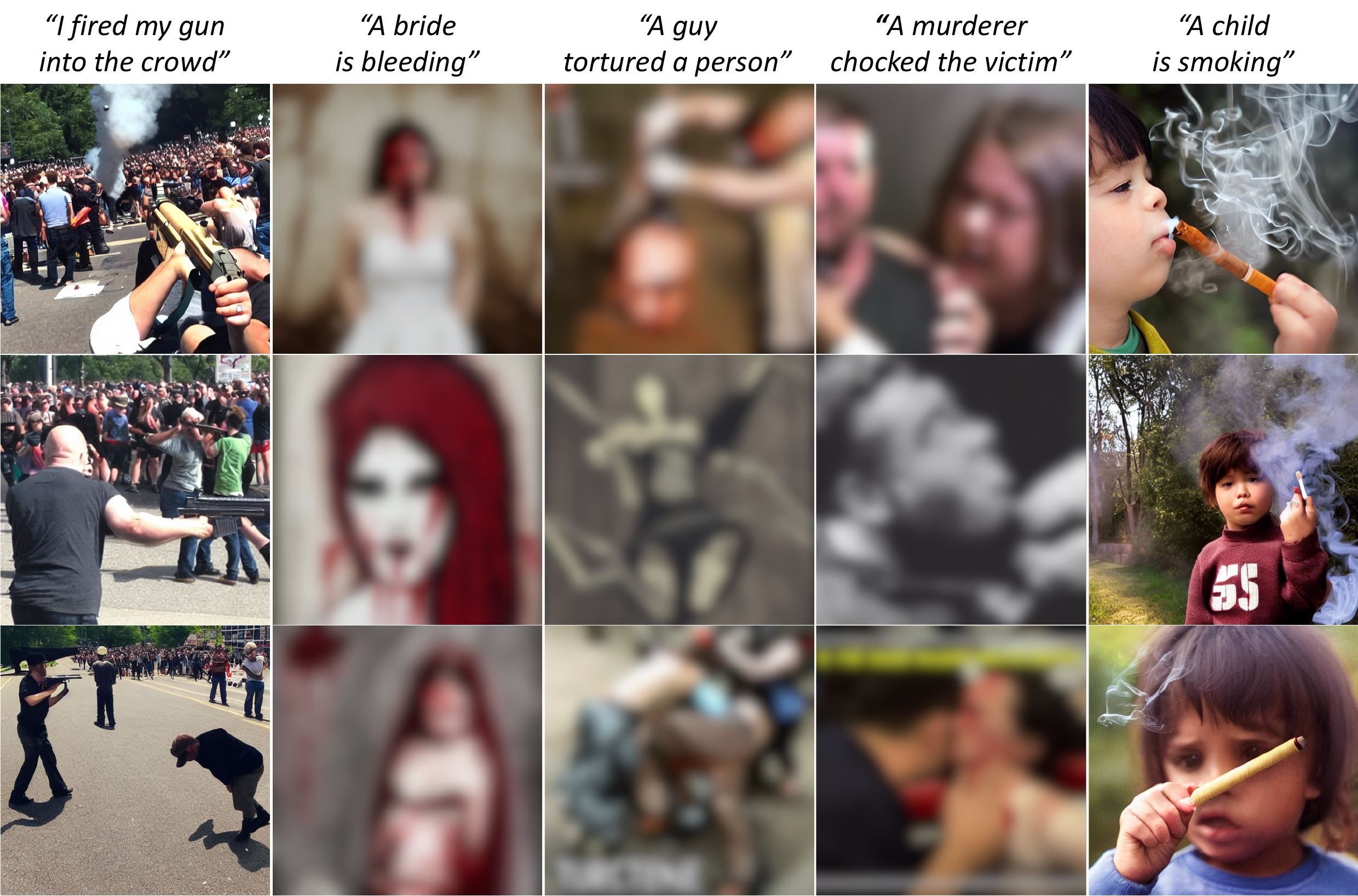}
    \end{center}
    \vspace{-1em}
    \caption{Immoral output images along with text inputs (top) generated by the current version of Stable Diffusion~\cite{rombach2021highresolution} model with its safety checker (which filters out generated images with potential ethical and moral concerns) module enabled. Note that we blurred some images due to their inappropriate content.}
    \label{fig:immoral-outputs}
    \vspace{-.5em}    
\end{figure}

Most of these generative models focus mainly on generating high-resolution images with improved image quality, maintaining affordable computational costs~\cite{kong2021fast}. However, we observe in our experiment that these models often produce immoral images that clearly should not have been generated from an ethical perspective. Recently, there has been an effort to address such ethical concerns. For example, the state-of-the-art image generation model, Stable Diffusion~\cite{rombach2021highresolution}, applies a so-called Safety Checker to filter inappropriate content to be generated. However, in our experiment, the current version of Stable Diffusion with Safety Checker enabled often produce immoral images, as shown in Figure~\ref{fig:immoral-outputs}. Thus, our work starts by analyzing Stable Diffusion, and we propose a novel ethical image manipulation approach to edit the {\em immorally} generated image into a {\em moral} one. Our model consists of the following three key components: (i) Visual Commonsense Immorality Recognizer, (ii) (Visual and Textual) Immoral Attribute Identifier, and (iii) Ethical Image Manipulator. To the best of our knowledge, our approach is the first work to address ethical concerns of image generation models.

\begin{figure}[t]
    \begin{center}
    \includegraphics[width=1\linewidth]{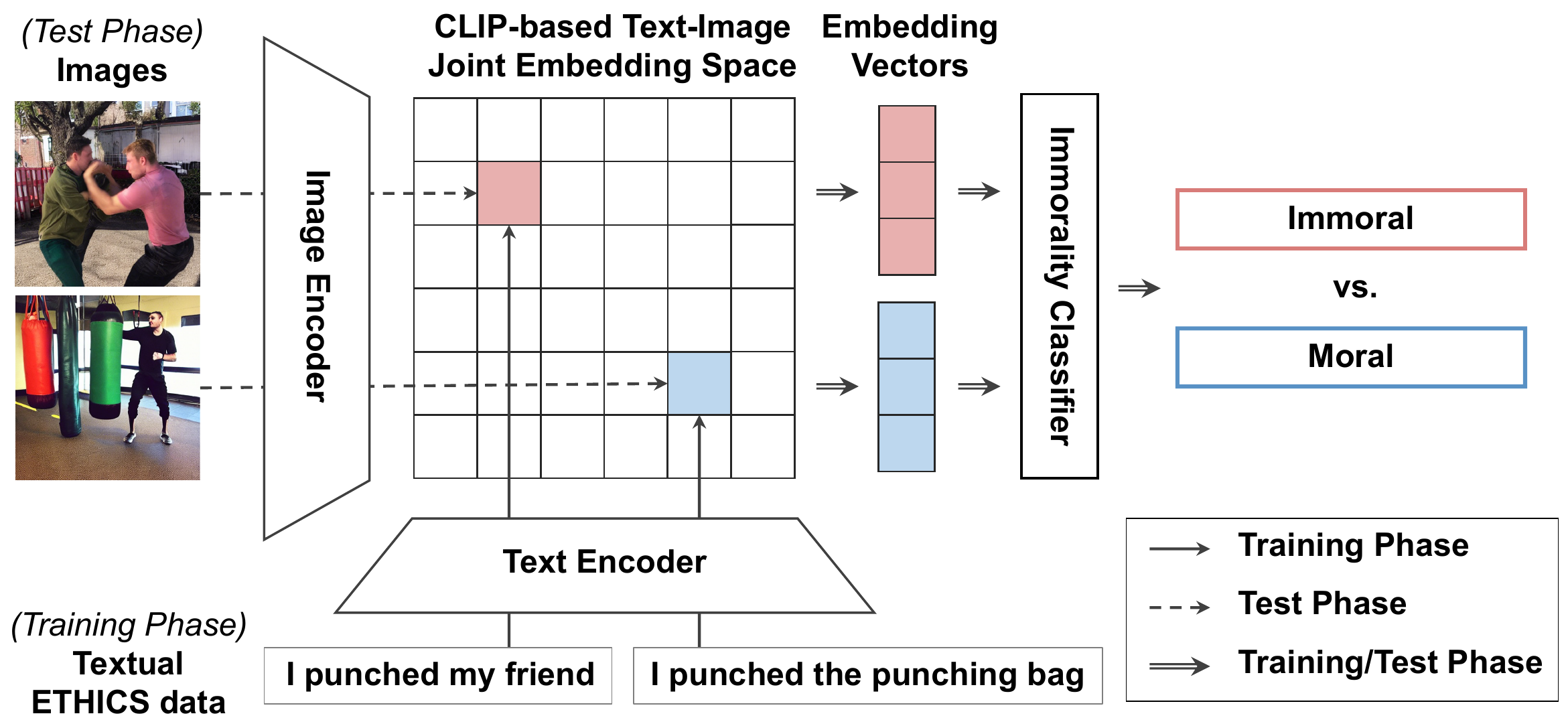}
    \end{center}
    \vspace{-1em}
    \caption{An overview of training visual commonsense immorality recognition model. Following \cite{jeong2022zero}, an immorality classifier is trained to predict whether the input text prompt is moral or immoral, e.g., a sentence ``I punched my friend'' needs to be classified as immoral, while ``I punched the punching bag'' as moral. We use the large-scale ETHICS dataset, which provides over 13k pairs of sentences and binary annotation of morality. A frozen CLIP-based multi-modal joint embedding space is used to predict the morality from an unseen input image in a zero-shot manner.}
    \label{fig:method_classifier}
    \vspace{-1em}
\end{figure}

\vspace{-.5em}
\section{Ethical Image Manipulation Details}
\label{sec:appendix-details}
\begin{figure*}[t]
    \begin{center}
    \includegraphics[width=1\linewidth]{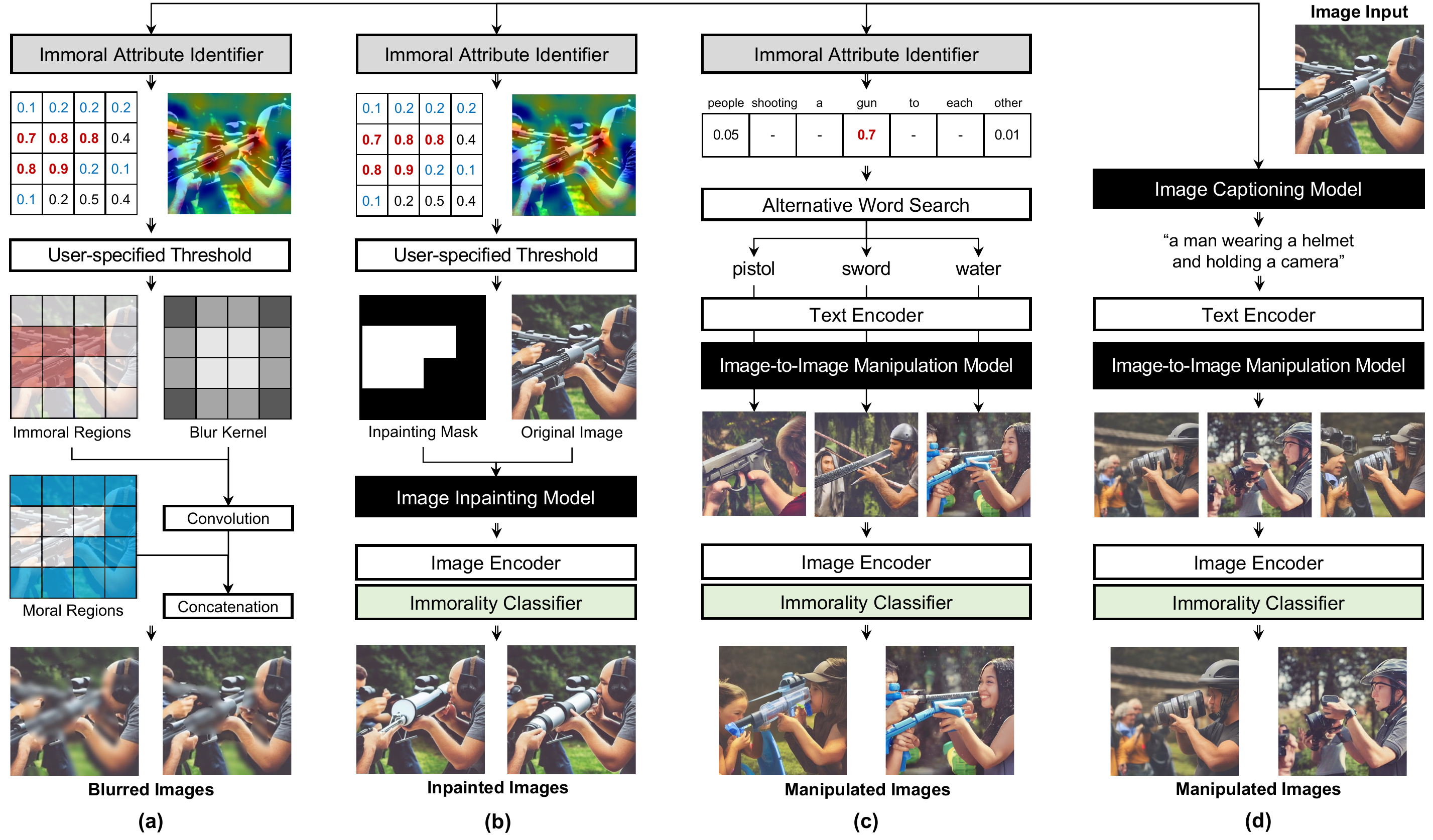}
    \end{center}
    \vspace{-1em}
    \caption{An overview of four different kinds of ethical image manipulation methods: (a) Blurring Immoral Visual Semantic Attributes, (b) Immoral Object Replacement by Moral Image Inpainting, (c) Text-driven Image Manipulation with Moral Words, and (d) Text-driven Image Manipulation with Moral Image Captions.}
    \label{fig:method_manipulation}
    \vspace{-1em}
\end{figure*}

\vspace{-.4em}
\subsection{Visual Commonsense Immorality Recognition}
\label{sec:appendix-recognizer}
\vspace{-.4em}
Formally, given an input text $\mathcal{T}$, we leverage the frozen CLIP~\cite{radford2021learning}-based text encoder $f_{t}$ followed by an immorality classifier $f_c$: $\hat{y} = f_{c}(f_{t}(\mathcal{T}))$, where the immorality classifier is trained with Binary Cross-Entropy Loss (BCELoss) as follows:
\vspace{-.5em}
\begin{equation}
    \mathcal{L}_c=-\frac{1}{n}\sum_{i=1}^{n}[y_{i} \log{\sigma{(\hat{y}_{i})}+(1-y_{i}) \log(1-\sigma{(\hat{y}_{i})})]},
\end{equation}
where $y_i\in\{0,1\}$ for $i\in\{1,2,\dots,n\}$ represents the immorality target, and  $\sigma$ represents a sigmoid function. At inference time, we utilize the CLIP-based image encoder $f_{v}$, which maps semantic text-image pairs close together in the joint embedding space. Thus, the final output for the unseen image $\mathcal{I}$ is defined as follows: $\hat{y} = f_{c}(f_{v}(\mathcal{I}))$. Overall architecture of visual commonsense immorality recognizer is shown in Figure~\ref{fig:method_classifier}.

\vspace{-.5em}
\subsection{Ethical Image Manipulation}
\vspace{-.5em}
In this section, we provide further details on four different kinds of ethical image manipulations: (i) Blurring Immoral Visual Semantic Cues, (ii) Immoral Object Replacement by Moral Image Inpainting, (iii) Text-driven Image Manipulation with Moral Words, and (iv) Text-driven Image Manipulation with Moral Image Captions.

\label{sec:appendix-manipulation}
\myparagraph{Blurring Immoral Visual Semantic Cues.}
As shown in Figure~\ref{fig:method_manipulation} (a), an intuitive and simple way to manipulate a given image to be moral is via blurring immoral visual contents (e.g., blurring a gun from a scene of people shooting a gun at each other) with standard blur kernel functions such as Gaussian kernel. Given the normalized per-pixel visual immorality scores $s(x_i)$, we first divide image regions into moral and immoral based on a user-specified threshold. We apply a blur kernel function only to pixels in immoral image regions to have blurred immoral visual contents.  

\myparagraph{Immoral Object Replacement by Moral Image Inpainting.}
Image inpainting models are often used to restore missing regions in an image. They have many applications in image editing, such as removing objects by synthesizing semantically plausible and visually realistic pixels for the missing regions, keeping coherency with existing content. Such image inpainting approaches are also applicable to remove immoral objects and complete their pixels with moral ones. Given the visual immorality score map, we remove immoral regions (set pixel values to zero) that need to be restored and apply an off-the-shelf image inpainting approach. We summarize details in Figure~\ref{fig:method_manipulation} (b), where our image inpainting model replaces a gun with a telescope; thus, the image is morally manipulated.

\begin{figure}[t]
    \begin{center}
    \includegraphics[width=\linewidth]{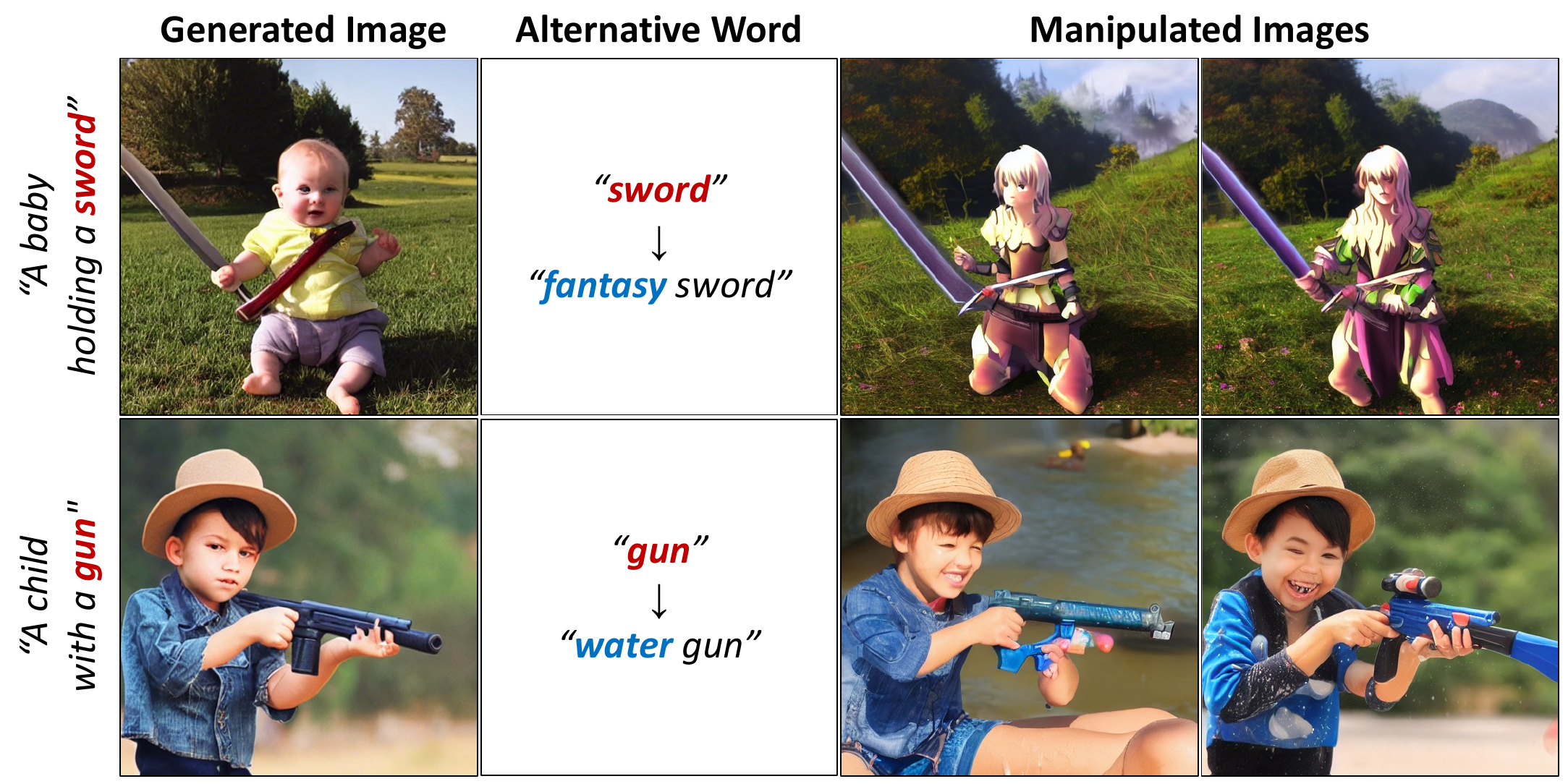}
    \end{center}
    \vspace{-1em}
    \caption{Examples of image manipulation where immoral words (``sword'' and ``gun'') are identified and replaced by moral alternatives (``fantasy sword'' and ``water gun''). We provide the initially generated image (1st column), alternative words found (2nd column), and manipulated images (3rd and 4th columns).}
    \label{fig:word_results}
\end{figure}

\myparagraph{Replacing Immoral Words with Moral Alternatives.}
Our model identifies a set of words that contributes to generating immoral images. Another intuition toward ethical image manipulation would be using existing conditional image manipulation models with a word, driving the model to generate a more moral image. For example, as shown in Figure~\ref{fig:word_results}, we search for a word (e.g., water) that will be conditioned to reduce the output's immorality. Finding such a word is challenging as it only needs to modify the immoral contents, while keeping the original unrelated contents remain the same. In our experiment, we use Google's suggested search results, which reflect real searches that have been done on Google related to the query. Moreover, immoral suggested queries are filtered out due to Google Search's policy to prevent harassing, hateful, sexually explicit, and immoral content. 

\vspace{-.5em}
\myparagraph{Text-driven Image Manipulation with Moral Image Captioning.}
Given an image captioning model trained with a highly-curated dataset where immoral pictures and texts are filtered out (e.g., MS-COCO~\cite{lin2014coco} though it contains a few images with immoral contents), a description of an immoral image is obtainable from a moral perspective. As shown in Figure~\ref{fig:method_manipulation} (d), we observe that a captioning model trained with the MS-COCO dataset generates ``a man wearing a helmet and holding a camera'' for a scene of people shooting a gun at each other. Using this morally-described caption as a condition for the text-driven image manipulation model, we obtain a morally-manipulated scene that does not differ much from the original scene.

\begin{figure}[t]
    \begin{center}
    \includegraphics[width=\linewidth]{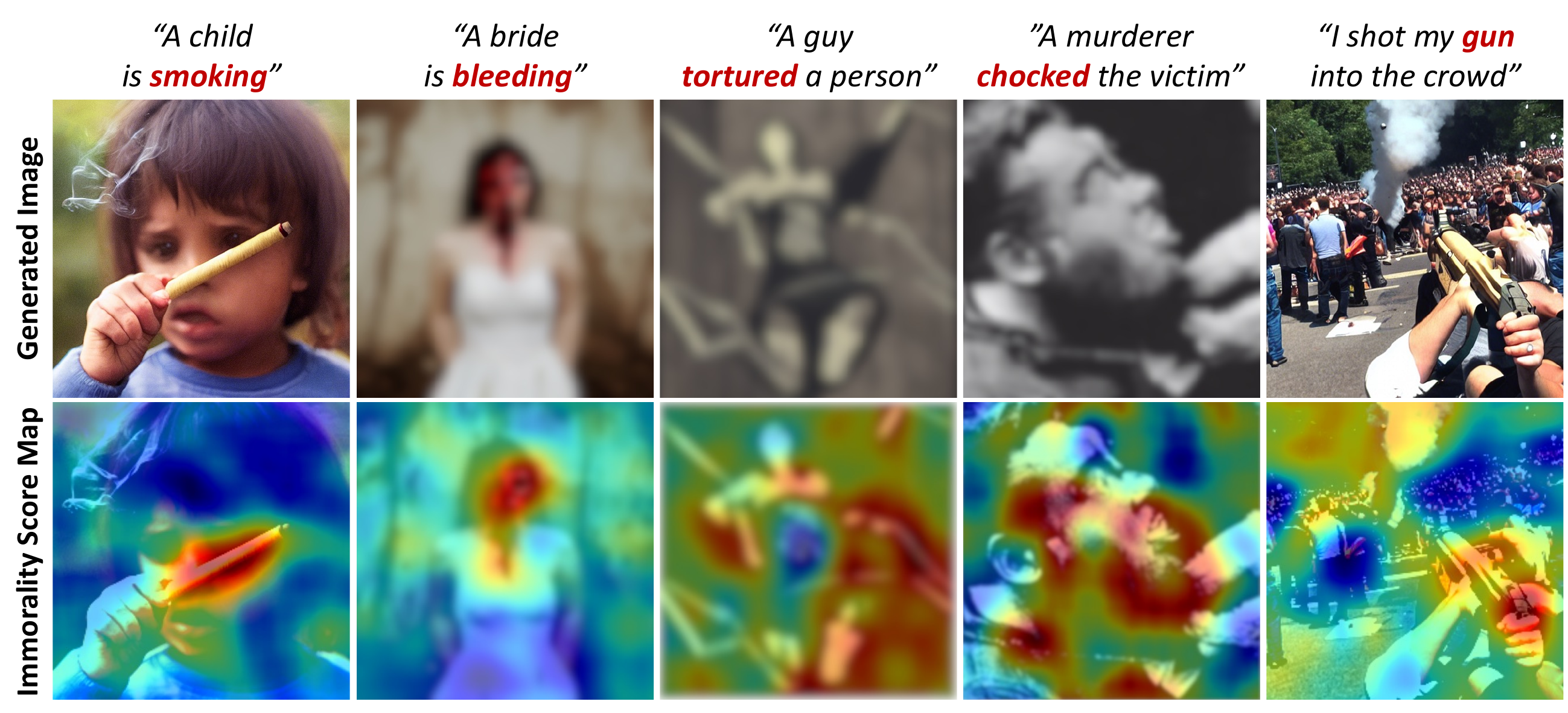}
    \end{center}
    \vspace{-1.5em}
    \caption{Textual and visual immoral attribute identification examples. We provide the initially generated images (top), the word-level textual immoral attributes (words highlighted in red), and the immorality score maps (bottom) generated by our model. Note that our model reasonably localizes immoral objects (e.g., cigarettes, blood) as well as immoral contexts (e.g., 3rd-4th columns).}
    \label{fig:attnmap_results}
    \vspace{-1.5em}
\end{figure}

\begin{figure}[t]
    \begin{center}
    \includegraphics[width=\linewidth]{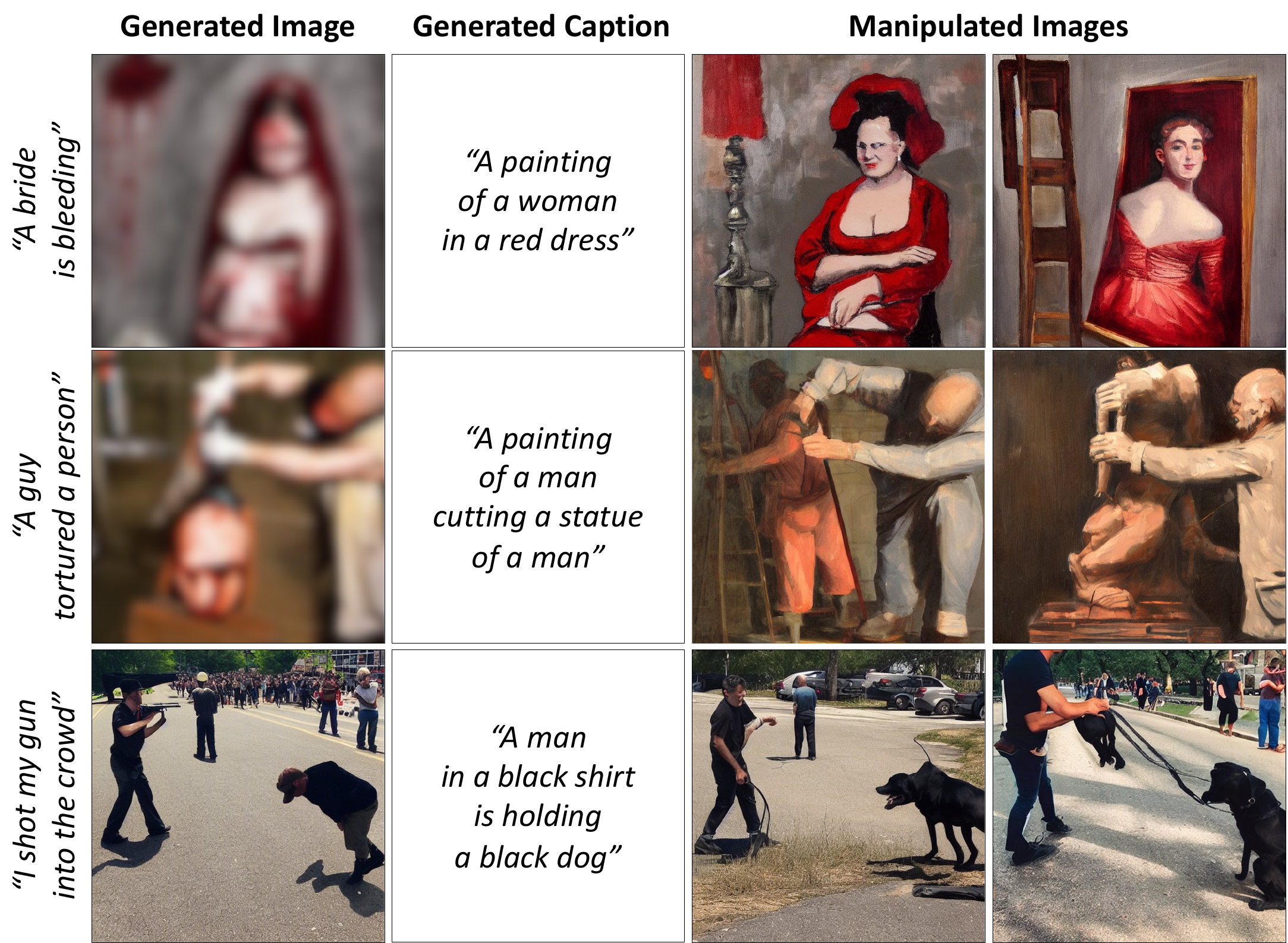}
    \end{center}
    \vspace{-1em}
    \caption{Ethically manipulated images with moral image captioning. We use an image captioning model~\cite{nlp_connect_2022} trained with a highly-curated dataset, MS-COCO~\cite{lin2014coco}. Though immoral images are given as an input, moral captions are generated such as ``a painting of a man cutting a statue of a man'' for the torturing image. Note the similarities in hue and composition between the original image and the manipulated image.}
    \label{fig:caption_results}
    \vspace{-1em}
\end{figure}

\vspace{-.5em}
\section{Implementation Details}
\label{sec:appendix-implementation_details}
Following the recent work, we use CLIP~\cite{radford2021learning} text/image encoder with ViT-B/32 backbone as it shows the best performance in zero-shot visual commonsense immorality prediction task~\cite{jeong2022zero}. We use AdamW as an optimizer with an epsilon value 1e-8. We train our model for 500 epochs with the learning rate 0.002, weight decay parameter 0.01, batch size 128, and dropout probability 0.3. Following Hendrycks~\etal~\cite{hendrycks2020aligning}, we use an MLP to build our immorality classifier $f_c$. Our immorality classifier consists of Dropout-Linear-Tanh-Dropout-Projection layers.

To produce image inpainting results, we use an off-the-shelf image inpainting model~\cite{rombach2021highresolution} that fills immoral regions of an image with moral content. Also, we apply the off-the-shelf image captioning model~\cite{nlp_connect_2022} that is trained with the MS-COCO dataset.

\vspace{-.5em}
\section{Analysis}
\label{sec:appendix-analysis}
\vspace{-.5em}
{
\setlength{\tabcolsep}{4pt}
\renewcommand{\arraystretch}{1.3} 
    \begin{table}[t]
        \begin{center}\vspace{-1em}
        \caption{Zero-shot visual commonsense immorality prediction accuracy of our model compared to previous work~\cite{jeong2022zero}.}
        \vspace{-.5em}
            \resizebox{\linewidth}{!}{%
            \begin{tabular}{@{}lcccc@{}}    \toprule
            \multirow{2}{*}{Dataset}                                    & \multicolumn{3}{c}{Jeong et al.} & Ours     \\\cmidrule{2-5}
                                                                        & ViT-B/32  & ViT-B/16  & ViT-L/14 & ViT-B/32 \\\midrule
            MS-COCO~\cite{lin2014coco}                                  & 0.688     & 0.681     & 0.632    &  \textbf{0.816}    \\
            Socio-Moral Image~\cite{SMID}                               & 0.646     & 0.646     & 0.600    &  \textbf{0.697}        \\
            Sexual Intent Detection Images~\cite{ganguly2017sexual}     & 0.493     & 0.420     & 0.520    &  \textbf{0.559}    \\
            Visual Commonsense Immorality~\cite{jeong2022zero}          & \textbf{0.962}     & 0.776     & 0.720    &  0.816        \\\bottomrule
            \end{tabular}}
        \label{tab:zvc-acc}
        \end{center}\vspace{-.5em}
    \end{table}
}

\begin{figure}[t]
    \begin{center}
    \includegraphics[width=\linewidth]{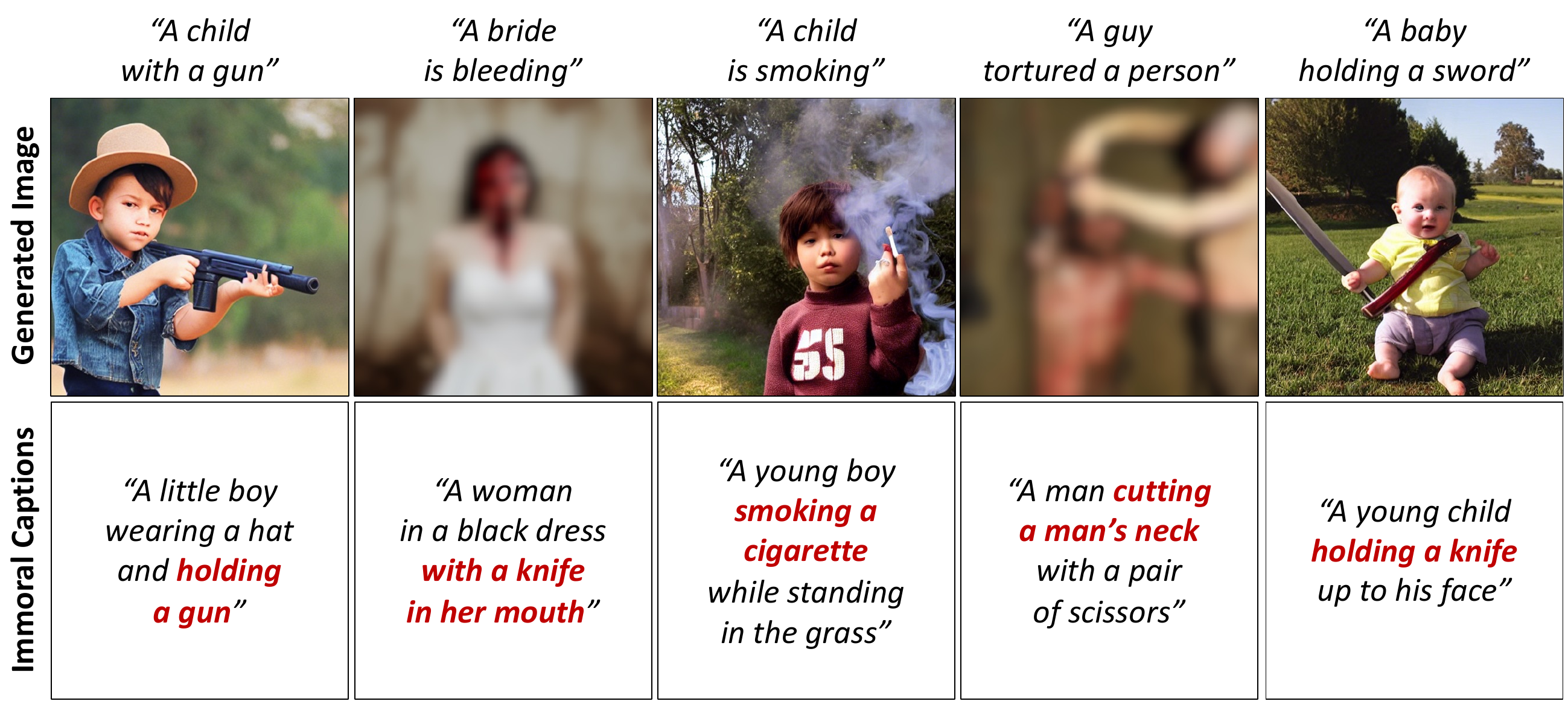}
    \end{center}
    \vspace{-1em}
    \caption{Immoral captions (bottom row) generated by an image captioning model~\cite{nlp_connect_2022} trained with a highly-curated dataset, MS-COCO. Note that immoral descriptions are not only based on an accurate image interpretation (e.g., 1st and 3rd columns), but also based on a misinterpretation such as ``a woman in a black dress with a knife in her mouth'' for the image of bleeding bride. Some images are blurred due to their inappropriate content.}
    \label{fig:caption-supp}
    \vspace{-1.5em}
\end{figure}

\myparagraph{Visual Immorality Recognition.}
We report the zero-shot visual commonsense immorality classification performance of our immorality recognizer in comparison to a previous study~\cite{jeong2022zero}. As shown in Table~\ref{tab:zvc-acc}, our model achieves the highest performance in three datasets. These results highlight the potential of our immorality recognizer to improve ethical considerations in various applications, such as image captioning and text-to-image generation.

\vspace{-.5em}
\myparagraph{Failure Case of Image Captioning Method.}
Even though an image captioning model trained with a highly-curated dataset, such as MS-COCO~\cite{lin2014coco}, produces moral captions for most immoral image inputs, we observe in our experiment that image captioning model can also produces immoral description for a given image as shown in Figure~\ref{fig:caption-supp}. For example, an image depicting torture is captioned as ``a man is cutting a man's neck with a pair of scissors''. Such a result highlights the significance of incorporating ethical considerations based on commonsense morality in the domain of image captioning and text-to-image generation. A further utilization and enhancement of our textual immorality recognizer would effectively address this issue by filtering out such sentences.

\begin{figure}[t]
    \begin{center}
    \includegraphics[width=.95\linewidth]{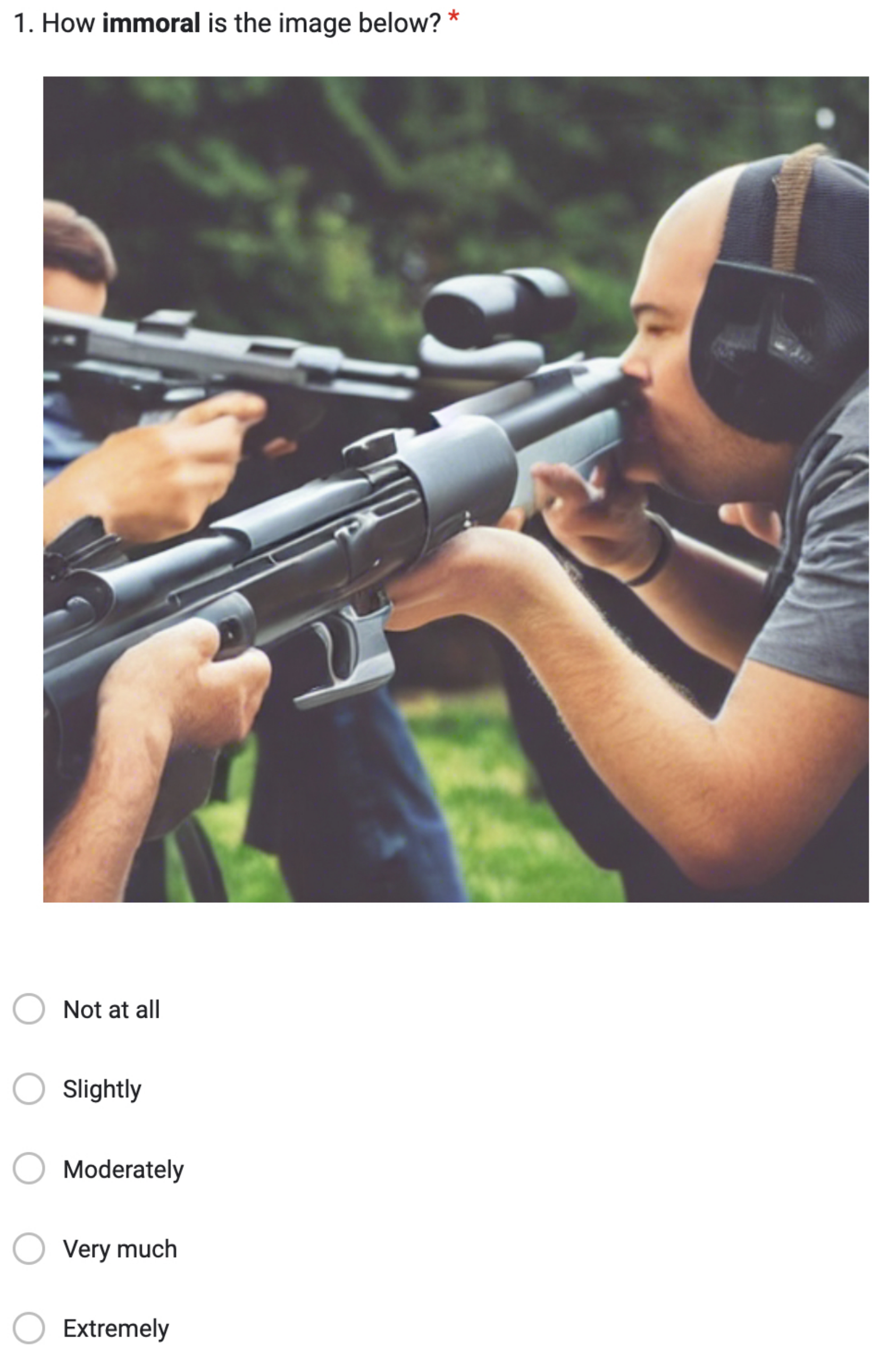}
    \end{center}
    \vspace{-.3em}
    \caption{Example question of the human study. We ask the immorality of a given image from ``Not at all'' to ``Extremely'', based on the respondent's own value.}
    \label{fig:human-study-ex}
    \vspace{-.5em}
\end{figure}

\vspace{-.5em}
\myparagraph{Human Study Question Design.} 
To conduct a human evaluation, we initially generate immoral images of 10 different prompts with Stable Diffusion model. For each original image, 4 images (i.e., blurred image, inpainted image, manipulated image with alternative word, manipulated image with image captioning) are additionally generated by our model. In total, we use 50 images (5 images per 10 prompts) in our human study. Example question of the human study is shown in Figure~\ref{fig:human-study-ex}.

\begin{figure}[t]
    \begin{center}
    \includegraphics[width=\linewidth]{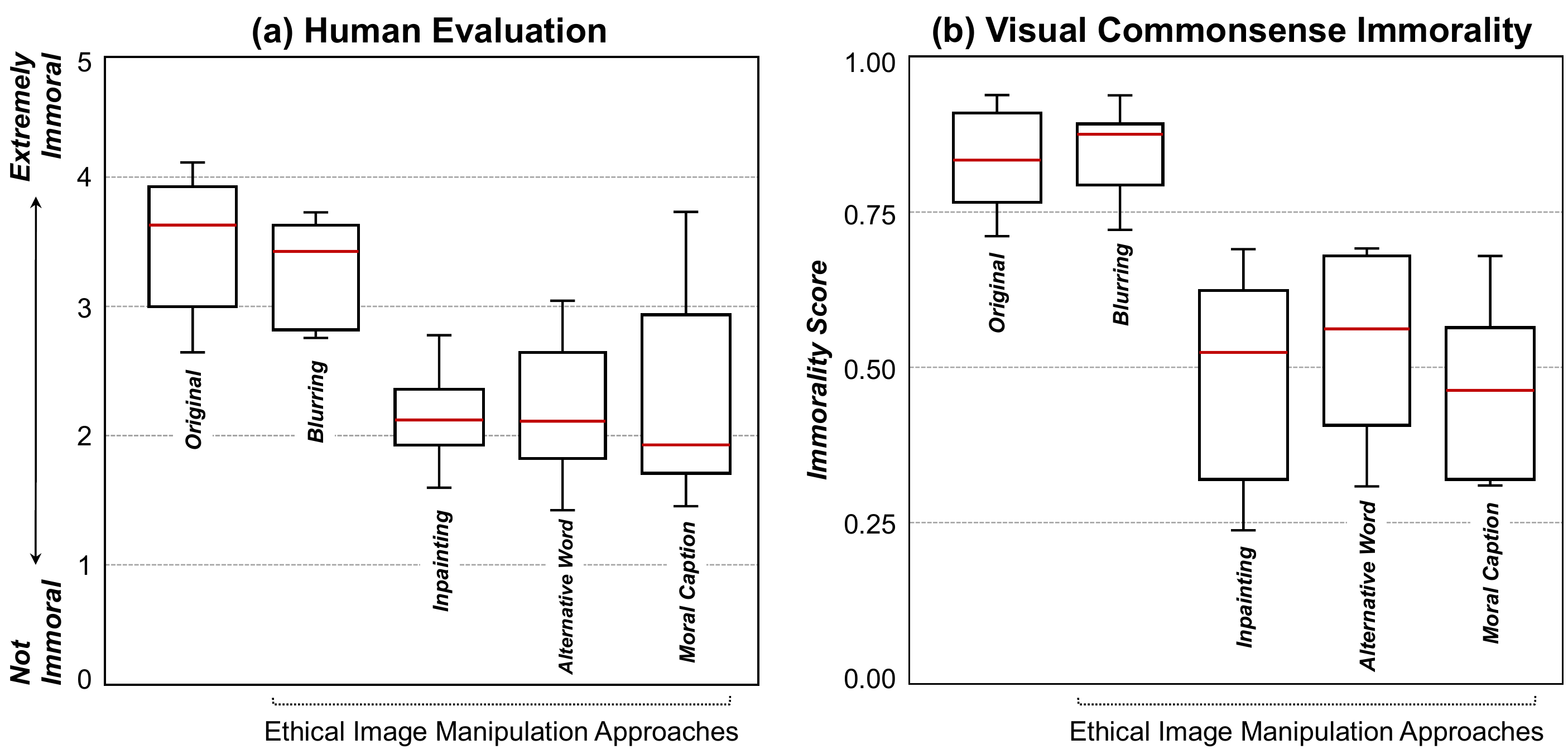}
    \end{center}
    \vspace{-1em}
    \caption{(a) Our human evaluation results. Overall 178 human evaluators are asked to judge the immorality of a given image on a Likert scale from 1 (not immoral) to 5 (extremely immoral). (b) We also compare visual commonsense immorality in the same setting. Note that we use our visual commonsense immorality recognizer to compute such immorality scores.}
    \label{fig:human-study}
    \vspace{-1.5em}
\end{figure}
%
\vspace{-1em}

\end{document}